\documentclass{article}
\pdfoutput=1
\usepackage{arxiv}

\def\BibTeX{{\rm B\kern-.05em{\sc i\kern-.025em b}\kern-.08em
    T\kern-.1667em\lower.7ex\hbox{E}\kern-.125emX}}

\usepackage[utf8x]{inputenc}
\usepackage{graphicx}
\usepackage{xcolor}
\usepackage{amsmath}
\usepackage{booktabs} 
\usepackage{caption}
\usepackage{subcaption}

\usepackage{multicol}
\usepackage{multirow}
\usepackage{fancyhdr}
\usepackage{url}
\usepackage{lipsum}
\usepackage{mathtools}
\usepackage{cuted}
\usepackage{enumitem}
\usepackage[linesnumbered, ruled, noline]{algorithm2e}
\usepackage{color}
\usepackage{amsthm}
\usepackage{amsfonts}
\usepackage{tabularx}

\newtheorem{definition}{Definition}

\title{Online Co-movement Pattern Prediction in Mobility Data}

\author{
  Andreas Tritsarolis, Eva Chondrodima, Panagiotis Tampakis and Aggelos Pikrakis\\
  Data Science Lab., Department of Informatics, University of Piraeus\\
  Piraeus, Greece \\
  \texttt{\{andrewt,evachon,ptampak,pikrakis\}@unipi.gr} \\
}

\begin{document}

\maketitle

\newcommand{\prob}{{Online Prediction of Co-movement Patterns}}

\begin{abstract}
Predictive analytics over mobility data are of great importance since they can assist an analyst to predict events, such as collisions, encounters, traffic jams, etc. A typical example of such analytics is future location prediction, where the goal is to predict the future location of a moving object, given a look-ahead time. What is even more challenging is being able to accurately predict collective behavioural patterns of movement, such as co-movement patterns. In this paper, we provide an accurate solution to the problem of \emph{\prob}. In more detail, we split the original problem into two sub-problems, namely \emph{Future Location Prediction} and \emph{Evolving Cluster Detection}. Furthermore, in order to be able to calculate the accuracy of our solution, we propose a co-movement pattern similarity measure, which facilitates us to match the predicted clusters with the actual ones. Finally, the accuracy of our solution is demonstrated experimentally over a real dataset from the maritime domain.
\end{abstract}



\maketitle

\section{Introduction}\label{sect:Introduction}
    
    The vast spread of GPS-enabled devices, such as smartphones, tablets and GPS trackers, has led to the production of large ammounts of mobility related data. 
    By nature, this kind of data are streaming and there are several application scenarios where the processing needs to take place in an online fashion. These properties have posed new challenges in terms of efficient storage, analytics and knowledge extraction out of such data. 
    One of these challenges is online cluster analysis, where the goal is to unveil hidden patterns of collective behavior from streaming trajectories, such as co-movement patterns~\cite{DBLP:journals/pvldb/FanZWT16, DBLP:journals/pvldb/ChenGFMJG19, DBLP:conf/sigmod/FangGPCMJ20, DBLP:conf/icde/HelmiK20, doi:10.1080/13658816.2020.1834562}. What is even more challenging is predictive analytics over mobility data, where the goal is to predict the future behaviour of moving objects, which can have a wide range of applications, such as predicting collisions, future encounters, traffic jams, etc. At an individual level, a typical and well-studied example of such analytics is future location prediction~\cite{DBLP:journals/is/TrasartiGMG17, s20185133, DBLP:conf/ssd/PetrouNTGKSPVGC19, DBLP:conf/pkdd/PetrouTGPT19}, where the goal is to predict the future location of a moving object, given a look-ahead time. However, prediction of future mobility behaviour at a collective level and more specifically \emph{\prob}, has not been addressed in the relevant literature yet. 
    
\begin{figure*}
        \centering
        \includegraphics[width=1\columnwidth]{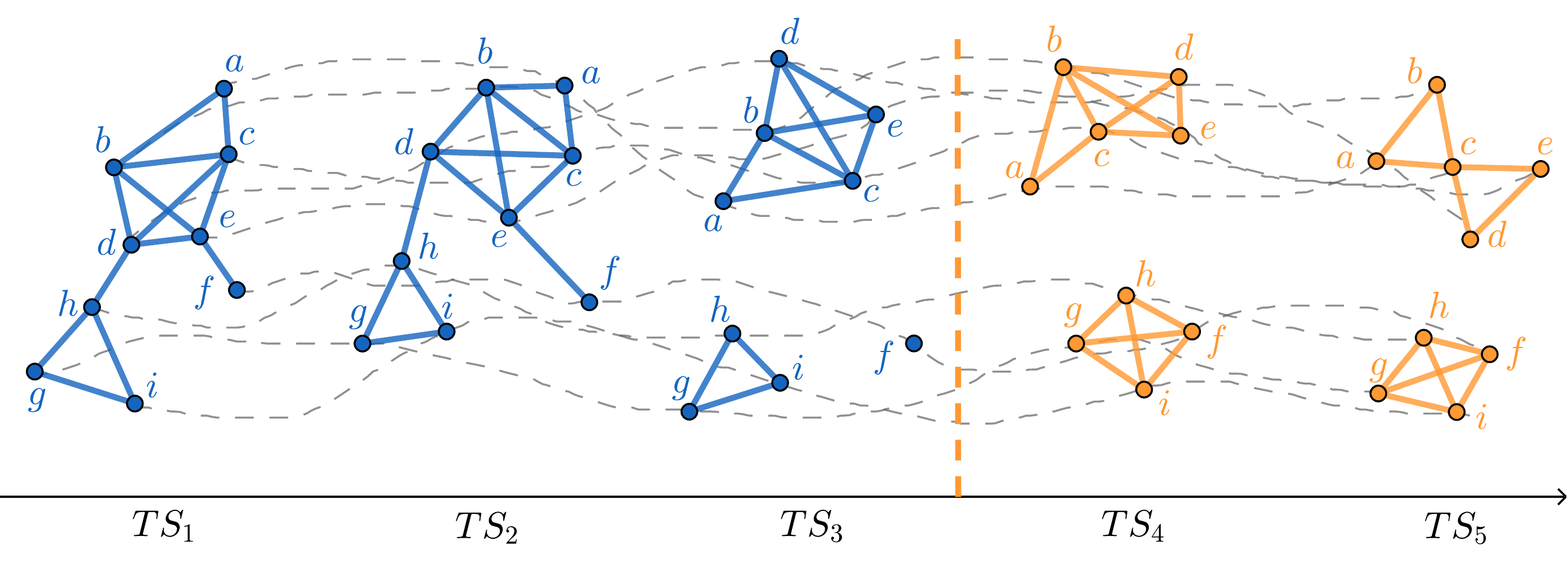}
        \caption{Predicting evolving clusters via (singular) trajectory prediction}
        \label{fig:Predicting_EC_Trajectory}
    \end{figure*}

    Concerning the definition of co-movement patterns, there are several approaches in the literature, such as~\cite{DBLP:journals/pvldb/FanZWT16, DBLP:conf/icde/HelmiK20, DBLP:journals/pvldb/ChenGFMJG19, DBLP:conf/sigmod/FangGPCMJ20}. However, all of the above are either offline and/or operate at predefined temporal snapshots that imply temporal alignment and uniform sampling, which is not realistic assumptions.  For this reason, we adopt the approach presented in~\cite{doi:10.1080/13658816.2020.1834562}, which, to the best of our knowledge, is the first online method for the discovery of co-movement patterns in mobility data that does not assume temporal alignment and uniform sampling. The goal in~\cite{doi:10.1080/13658816.2020.1834562} is to discover co-movement patterns, namely \emph{Evolving Clusters}, in an online fashion, by employning a graph-based representation. By doing so, the problem of co-movement pattern detection is transformed to identifying \emph{Maximal Cliques} (MCs) (for spherical clusters) or \emph{Maximal Connected Subgraphs} (MCSs) (for density-connected clusters). Figure~\ref{fig:Predicting_EC_Trajectory} illustrates such an example, where in blue we have the historical evolving clusters and in orange the predicted future ones. 
    
    Several mobility-related applications could benefit from such an operation. In the urban traffic domain, predicting co-movement patterns could assist in detecting future traffic jams which in turn can help the authorities take the appropriate measures (e.g. adjusting traffic lights) in order to avoid them. In the maritime domain, a typical problem is illegal transshipment, where groups of vessels move together "close" enough for some time duration and with low speed. It becomes obvious that predicting co-movement patterns could help in predicting illegal transshipment events. Finally, in large epidemic crisis, contact tracing is one of the tools to identify individuals that have been close to infected persons for some time duration. Being able to predict groups of individuals that will be close to each other for some duration can help avoid future contacts with possibly infected individuals.
    
    The problem of predicting the spatial properties of group patters has only been recently studied~\cite{DBLP:conf/gis/KannangaraXTHK20}. In more detail, the authors in~\cite{DBLP:conf/gis/KannangaraXTHK20} adopt a spherical definition of groups, where each group consists of moving objects that are confined within a radius $d$ and their goal is to predict the centroid of the groups at the next timeslice. However, this approach is offline and cannot be applied in an online scenario. Furthermore, the group definition adopted in~\cite{DBLP:conf/gis/KannangaraXTHK20} is rather limited, since the identify only spherical groups, as opposed to~\cite{doi:10.1080/13658816.2020.1834562} where both spherical and density-connected clusters can be identified. Finally, the authors in~\cite{DBLP:conf/gis/KannangaraXTHK20} predict only the centroids of the clusters and not the shape and the membership of each cluster. 
    
    Inspired by the above, the problem that we address in this paper is the \emph{\prob}. Informally, given a look-ahead time interval $\Delta t$, the goal is to predict the groups, i.e. their spatial shape (spherical or density-connected), temporal coverage and membership, after $\Delta t$ time. In more detail, we split the original problem into two sub-problems, namely \emph{Future Location Prediction} and \emph{Evolving Cluster Detection}. The problem of \emph{\prob} is quite challenging, since, apart from the inherent difficulty of predicting the future, we also need to define how the error between the actual and the predicted clusters will be measured. This further implies that a predicted cluster should be correctly matched with the corresponding actual cluster which is not a straightforward procedure. To the best of our knowledge, the problem of \emph{\prob}, has not been addressed in the literature yet.
    Our main contributions are the following: 
    \begin{itemize}
        \item We provide an accurate solution to the problem of \emph{\prob}.
        \item We propose a co-movement pattern similarity measure, which helps us ``match'' the predicted with the actual clusters.
        \item We perform an experimental study with a real dataset from the maritime domain, which verifes the accuracy of our proposed methodology.
    \end{itemize}
    
    The rest of the paper is organized as follows. Section~\ref{sect:RelatedWork} discusses related work. In Section~\ref{sect:ProblemDefinition}, we formally define the problem of \emph{\prob}. Subsequently, in Section~\ref{sect:Methodology} we propose our two-step methodology and in Section~\ref{sect:EvaluationMeasures}, we introduce a co-movement pattern similarity measure along with cluster ``matching'' algorithm. Section~\ref{sect:ExperimentalStudy}, presents our experimental findings and, finally, in Section~\ref{sect:ConclusionsAndFutureWork} we conclude the paper and discuss future extensions.

\section{Related Work}\label{sect:RelatedWork}

    The work performed in this paper is closely related to three topics, (a) trajectory clustering and more specifically co-movement pattern discovery, (b) future location prediction and (c) co-movement pattern prediction.
    
    \textbf{Co-movement patterns.} 
    One of the first approaches for identifying such collective mobility behavior is the so-called flock pattern~\cite{DBLP:journals/gis/LaubeIW05}, which identifies groups of at least $m$ objects that move within a disk of radius $r$ for at least $k$ consecutive timepoints.
    Inspired by this, several related works followed, such as moving clusters~\cite{DBLP:conf/ssd/KalnisMB05}, convoys~\cite{DBLP:journals/pvldb/JeungYZJS08}, swarms~\cite{DBLP:journals/pvldb/LiDHK10}, platoons~\cite{DBLP:journals/dke/LiBK15}, traveling companion~\cite{DBLP:conf/icde/TangZYHLHP12} and gathering pattern~\cite{DBLP:conf/icde/ZhengZYS13}. Even though all of these approaches provide explicit definitions of several mined patterns, their main limitation is that they search for specific collective behaviors, defined by respective parameters.
    An approach that defines a new generalized mobility pattern is presented in~\cite{DBLP:journals/pvldb/FanZWT16}. In more detail, the general co-movement pattern (GCMP), is proposed, which includes \textit{Temporal Replication} and \textit{Parallel Mining}, a method that, as suggested by its name, splits a data snapshot spatially and replicates data when necessary to ensure full coverage, and \textit{Star Partitioning} and \textit{ApRiori Enumerator}, a technique that uses graph pruning in order to avoid the data replication that takes place in the previous method. 
    In~\cite{DBLP:conf/icde/HelmiK20}, the authors propose a frequent co-movement pattern (f-CoMP) definition for discovering patterns at multiple spatial scales, also exploiting the overall shape of the objects’ trajectories, while at the same time it relaxes the temporal and spatial constraints of the seminal works (i.e. Flocks, Convoys, etc.) in order to discover more interesting patterns.
    The authors in~\cite{DBLP:journals/pvldb/ChenGFMJG19, DBLP:conf/sigmod/FangGPCMJ20}, propose a two-phase online distributed co-movement pattern detection framework, which includes the clustering and the pattern enumeration phase, respectively. During the clustering phase for timestamp $t_s$, the snapshot $S_t$ is clustered using Range-Join and DBSCAN.
    
    Another line of research, tries to discover groups of either entire or portions of trajectories considering their routes. There are several approaches whose goal is to group whole trajectories, including T-OPTICS~\cite{DBLP:journals/jiis/NanniP06, DBLP:journals/tsas/PelekisSTT16}, that incorporates a trajectory similarity function into the OPTICS algorithm.
    However, discovering clusters of complete trajectories can overlook significant patterns that might exist only for portions of their lifespan. To deal with this, another line of research has emerged, that of \emph{Subtrajectory Clustering}\cite{DBLP:conf/edbt/PelekisTVPT17, DBLP:journals/datamine/PelekisTVDT17, DBLP:conf/icde/TampakisPAAFT18, DBLP:conf/bigdataconf/TampakisPDT19}, where the goal is to partition a trajectory into subtrajectories, whenever the density or the composition and its neighbourhood changes ``significantly'', then form groups of similar ones, while, at the same time, separate the ones that fit into no group, called outliers.
    
    Another perspective into co-movement pattern discovery, is to reduce cluster types into graph properties and view them as such. In \cite{DBLP:conf/pkdd/TheodoropoulosT19, doi:10.1080/13658816.2020.1834562}, the authors propose a 
    novel co-movement pattern definition, called \textit{evolving clusters}, that unifies the definitions of flocks and convoys and reduces them to Maximal Cliques (MC), and Connected Components (MCS), respectively. In addition, the authors propose an online algorithm, that discovers several evolving cluster types simultaneously in real time using Apache Kafka\textsuperscript{\textregistered}, without assuming temporal alignment, in constrast to the seminal works (i.e. flocks, convoys).
    
    In the proposed predictive model, we will use the definition of \textit{evolving clusters}~\cite{doi:10.1080/13658816.2020.1834562} for co-movement pattern discovery. The reason why is this the most appropriate, is that we can predict the course of several pattern types at the same time, without the need to call several other algorithms, therefore adding redundant computational complexity.

    \textbf{Future Location Prediction.} 
    The fact that the Future Location Prediction (\emph{FLP}) problem has been extensivelly studied brings up its importance and applicability in a wide range of applications.  
    Towards tackling the \emph{FLP} problem, on line of work includes efforts that take advantage of historical movement patterns in order to predict the future location.
    Such an approach is presented in~\cite{DBLP:journals/is/TrasartiGMG17}, where the authors propose MyWay, a hybrid, pattern-based approach that utilizes individual patterns when available, and when not, collective ones, in order to provide more accurate predictions and increase the predictive ability of the system.
    In another effort, the authors in \cite{DBLP:conf/ssd/PetrouNTGKSPVGC19, DBLP:conf/pkdd/PetrouTGPT19} utilize the work done by \cite{DBLP:conf/bigdataconf/TampakisPDT19} on distributed subtrajectory clustering in order to be able to extract individual subtrajectory patterns from big mobility data. These patterns are subsequently utilized in order to predict the future location of the moving objects in parallel.
    
    A different way of addressing the \emph{FLP} problem includes machine learning approaches. 
    
    Recurrent Neural Network (RNN) -based models \cite{Rumelhart1986LearningRB} constitute a popular method for trajectory prediction due to their powerful ability to fit complex functions, along with their ability of adjusting the dynamic behavior as well as capturing the causality relationships across sequences. However, research in the maritime domain is limited regarding vessel trajectory prediction and Gated Recurrent Units (GRU) \cite{cho2014GRU} models, which constitute the newer generation of RNN. 
    
    Suo et.al. \cite{s20185133} 
    presented a GRU model to predict vessel trajectories based on a) the Density-Based Spatial Clustering of Applications with Noise (DBSCAN) algorithm to derive main trajectories and, b) a symmetric segmented-path distance approach to eliminate the influence of a large number of redundant data and to optimize incoming trajectories. Ground truth data from AIS raw data in the port of Zhangzhou, China were used to train and verify the validity of the proposed model.
    
    Liu et.al. \cite{8970798} 
    proposed a trajectory classifier called Spatio-Temporal GRU to model the spatio-temporal correlations and irregular temporal intervals prevalently presented in spatio-temporal trajectories. Particularly, a segmented convolutional weight mechanism was proposed to capture short-term local spatial correlations in trajectories along with an additional temporal gate to control the information flow related to the temporal interval information. 
    
    Wang et.al. \cite{9270462}
    aiming at predicting the movement trend of vessels in the crowded port water of Tianjin port, proposed a vessel berthing trajectory prediction model based on bidirectional GRU (Bi-GRU) and cubic spline interpolation.

    \textbf{Co-movement pattern prediction.}
    The most similar work to ours has only been recently presented in~\cite{DBLP:conf/gis/KannangaraXTHK20}. More specifically, the authors in~\cite{DBLP:conf/gis/KannangaraXTHK20}, divide time into time slices of fixed step size and adopt a spherical definition of groups, where each group consists of moving objects that are confined within a radius $d$ and their goal is to predict the centroid of the groups at the next timeslice. However, this approach is offline and cannot be applied in an online scenario. Furthermore, the group definition adopted in~\cite{DBLP:conf/gis/KannangaraXTHK20} is rather limited, since the identify only spherical groups, as opposed to~\cite{doi:10.1080/13658816.2020.1834562} where both spherical and density-connected clusters can be identified. Finally, the authors in~\cite{DBLP:conf/gis/KannangaraXTHK20} predict only the centroids of the clusters and not the shape and the membership of each cluster.

\section{Problem Definition} \label{sect:ProblemDefinition}
    

    As already mentioned, we divide the problem into two sub-problems, namely \emph{Future Location Prediction} and \emph{Evolving Clusters Detection}. Before proceeding to the actual formulation of the problem, let us provide some preliminary definitions.
    \begin{definition}
        (Trajectory) A trajectory $T = \{p_1, \dots p_n\}$ is considered as a sequence of timestamped locations, where $n$ is the latest reported position of $T$. Further, $p_i = \{x_i, y_i, t_i\}$, with $1 \leq i \leq n$.
    \end{definition}
    

    \begin{definition}\label{def:ProblemFormulationforNN} 
        (Future Location Prediction). Given an input dataset $D = \{T_1, \dots, T_{|D|}\}$ of trajectories and a time interval $\Delta t$, our goal is $\forall T_i \in D$ to predict $p^{i}_{pred} = \{x^{i}_{pred}, y^{i}_{pred}\}$ at timestamp $t^{i}_{pred} = t^{i}_{n}+\Delta t$.
        
    \end{definition}
    
    An informal definition regarding \textit{group patterns} could be: ``a large enough number of objects moving close enough to each other, in space and time, for some time duration''. As already mentioned, in this paper we adopt the definition provided in~\cite{doi:10.1080/13658816.2020.1834562}.
    
    \begin{definition}\label{def:GroupPattern} 
        (Evolving Cluster). Given: a set $D$ of trajectories, a minimum cardinality threshold $c$, a maximum distance threshold $\theta$, and a minimum time duration threshold $d$, an Evolving Cluster $\langle C, t_{start}, t_{end}, tp \rangle$ is a subset $C \in D$ of the moving objects’ population, $\lvert C \rvert \geq c$, which appeared at time point $t_{start}$ and remained alive until time point $t_{end}$ (with $t_{end} - t_{start} \geq d$) during the lifetime  $\lbrack t_{start}, t_{end} \rbrack$ of which the participating moving objects were spatially connected with respect to distance $\theta$ and cluster type $tp$.
        
    \end{definition}

    \begin{definition}\label{def:GroupPatternTrajectoryPredictionOnline} 
        (Group Pattern Prediction Online). Given: a set $D$ of trajectories, $G$ of co-movement patterns up to time-slice $TS^{now}$ and a lookahead threshold $\Delta t$, we aim to predict all the valid co-movement patterns $G' \in (TS^{now}, TS^{now} + \Delta t]$.
        
    \end{definition}

    Figure~\ref{fig:Predicting_EC_Trajectory} provides an illustration of Definition~\ref{def:GroupPatternTrajectoryPredictionOnline}. More specifically, we know the movement of nine objects from $TS_1$ until $TS_3$ and via EvolvingClusters with $c=3$ and $d=2$ that they form four evolving clusters $P_1 = \lbrace a,b,c,d,e,f,g,h,i \rbrace$, $P_2 = \lbrace a,b,c,d,e \rbrace$, $P_3 = \lbrace a,b,c \rbrace$, $P_4 = \lbrace b,c,d,e \rbrace$, $P_5 = \lbrace g,h,i \rbrace$. 
    Our goal is to predict their respective locations until $TS_5$. Running EvolvingClusters with the same parameters for the predicted time-slices, reveals us (with high probability) that $P_2, P_3, P_4, P_5$ will continue to exist as well as the creation of a new pattern $P_6 = \lbrace f,g,h,i \rbrace$.



\section{Methodology}\label{sect:Methodology}
    In this section we present the proposed solution to the problem of \emph{Online Prediction of Co-movement Patterns}, composed of two parts: a) the FLP method, and b) the Evolving Cluster Discovery algorithm. Also, an example is presented illustrating the approach operation.
    
    \begin{figure*}
        \includegraphics[width=1\columnwidth]{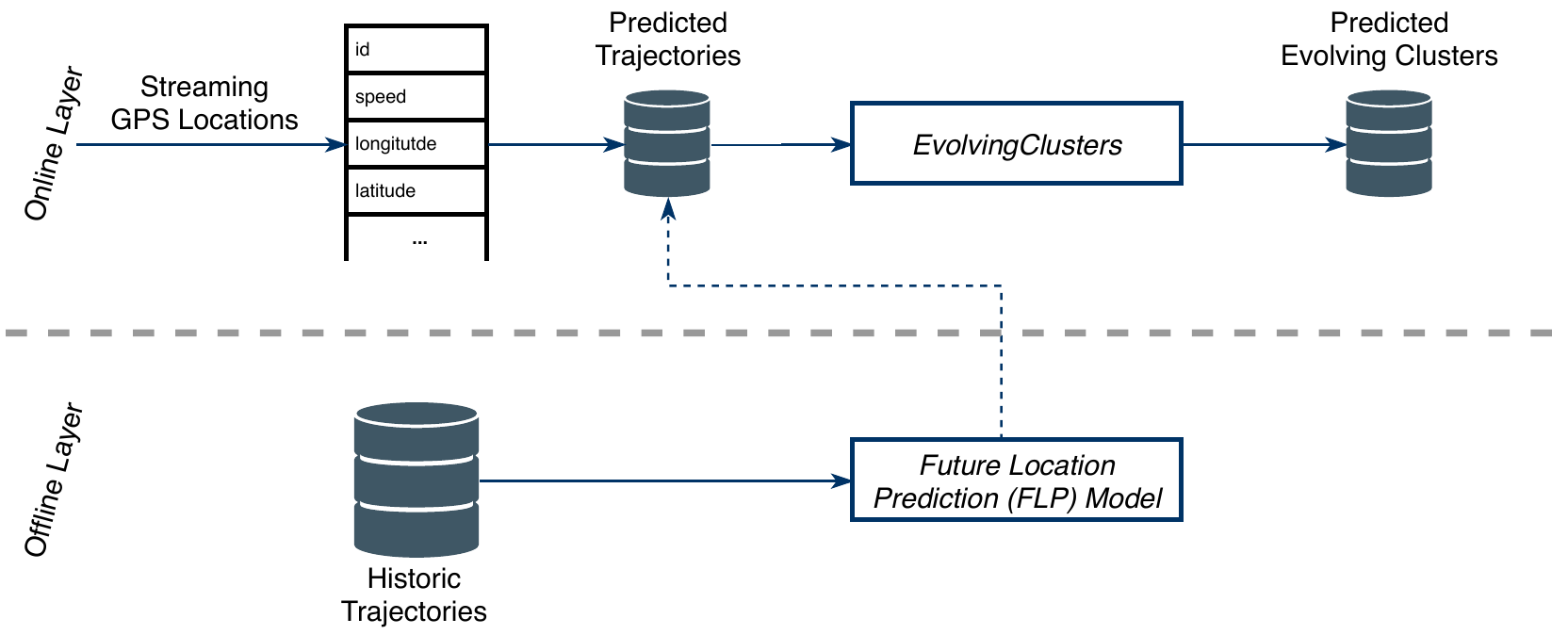}
        \captionsetup{justification=centering}
        
        \caption{Workflow for evolving clusters prediction via (singular) trajectory prediction}
        \label{fig:Predicting_EC_Trajectory_workflow}
    \end{figure*}
    
    \subsection{Overview}\label{subsect:Overview}
        Figure~\ref{fig:Predicting_EC_Trajectory_workflow} illustrates the architecture of our proposed methodology. First we split the problem of \emph{Online Prediction of Co-movement Patterns} into two parts, the FLP, and the Evolving Cluster Discovery. The FLP method is, also, divided to two parts: a) the FLP-offline part, where the training procedure of the model is taking place, and b) the FLP-online part, where the trained FLP model is applied to streaming GPS locations to predict the next objects' location.

        Thus, our proposed approach is further divided in the offline phase and the online one. Particularly, at the offline phase, we train our FLP model by using historic trajectories. Afterwards, at the online phase we receive the streaming GPS locations in order to use them to create a buffer for each moving object. Then, we use our trained FLP model to predict the next objects' location and apply EvolvingClusters to each produced time-slice.

    \subsection{Future Location Prediction}\label{subsect:FutureLocationPrediction}
        
        
        Trajectories can be considered as time sequence data \cite{8354239} and thus are suited to be treated with techniques that are capable of handling sequential data and/or time series \cite{08741193_TITS2020}. Over the past two decades, the research interest on forecasting time series has been moved to RNN-based models, with the GRU architecture being the newer generation of RNN, which has emerged as an effective technique for several difficult learning problems (including sequential or temporal data -based applications) \cite{8053243}. 
        Although, the most popular RNN-based architecture is the well-known Long Short-Term Memory (LSTM) \cite{Hochreiter1997}, GRU present some interesting advantages over the LSTM. More specifically, GRU are less complicated, easier to modify and faster to train. Also, GRU networks achieve better accuracy performance compared to LSTM models on trajectory prediction problems on various domains, such as on maritime \cite{s20185133}, on aviation \cite{9081618} and on land traffic \cite{8927604}. Hence, this work follows this direction and employs a GRU-based method.
        
        GRU includes internal mechanisms called gates that can regulate the flow of information. Particurlay, the GRU hidden layer include two gates, a reset gate which is used to decide how much past information to forget and an update gate which decides what information to throw away and what new information to add. We briefly state the update rules for the employed GRU layer. For more details, the interested reader is referred to the original publications \cite{cho2014GRU}. Also, details for the BPTT algorithm, which was employed for training the model, can be found in \cite{58337}.
        
        
        \begin{equation}
        \textbf{z}_k = \sigma({\textbf{W}}_{\tilde{\textbf{p}}z} \cdot \tilde{\textbf{p}}_{k} + {\textbf{W}}_{hz} \cdot {\textbf{h}}_{k-1} + {\textbf{b}}_z) 
        \end{equation}
        \begin{equation}
        \textbf{r}_k = \sigma({\textbf{W}}_{\tilde{\textbf{p}}r} \cdot \tilde{\textbf{p}}_{k} + {\textbf{W}}_{hr} \cdot {\textbf{h}}_{k-1} + {\textbf{b}}_r) 
        \end{equation}
        \begin{equation}
        \tilde{\textbf{h}}_k = \tanh({\textbf{W}}_{\tilde{\textbf{p}}h} \cdot \tilde{\textbf{p}}_{k} + {\textbf{W}}_{hh} \cdot (\textbf{r}_k *  {\textbf{h}}_{k-1}) + {\textbf{b}}_h)
        \end{equation}
        \begin{equation}
        \textbf{h}_k = \textbf{z}_k \odot \textbf{h}_{k-1} +(1-\textbf{z}_k) \odot \tilde{\textbf{h}}_k
        \end{equation}
        
        where $\textbf{z}$ and $\textbf{r}$ represent the update and reset gates, respectively, $\tilde{\textbf{h}}$ and $\textbf{h}$ represent the intermediate memory and output, respectively. Also, in these equations, the $\textbf{W}_{∗}$ variables are the weight matrices and the $\textbf{b}_{∗}$ variables are the biases. Moreover, $\tilde{\textbf{p}}$ represents the input, which is composed of the differences in space (longitude and latitude), the difference in time and the time horizon for which we want to predict the vessel’s position; the differences are computed between consecutive points of each vessel.
        
\begin{figure*}
            \centering
            \includegraphics[width=1\textwidth]{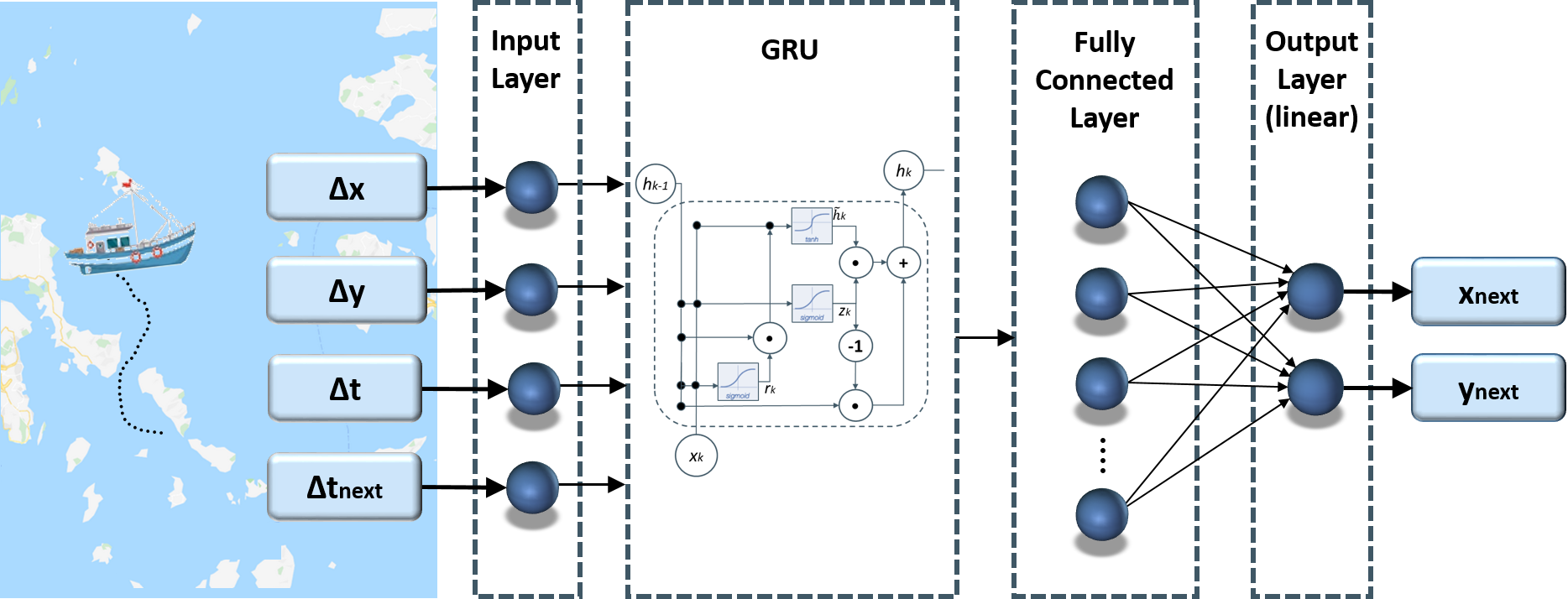}
            \caption{GRU-based neural network architecture.\label{fig:GRU}}
        \end{figure*}

        In this work, a GRU-based model is employed to solve the future location prediction problem. The proposed GRU-based network architecture is composed of the following layers: a) an input layer of four neurons, one for each input variable, b) a single GRU hidden layer composed of 150 neurons, c) a fully-connected hidden layer composed of 50 neurons, and d) an output layer of two neurons, one for each prediction coordinate (longitude and latitude).
        A schematic overview of the proposed network architecture is presented in Figure \ref{fig:GRU}. 
        Also, details for the Backward Propagation Through Time algorithm and for the Adam approach, which were employed for the NN learning purposes, can be found in \cite{Werbos1990} and \cite{Adam2015}, respectively. 
        
    \subsection{Evolving Clusters Discovery}\label{subsect:EvolvingClustersDiscovery}
        After getting the predicted locations for each moving object, we use \textit{EvolvingClusters} in order to finally present the predicted co-movement patterns. Because the sampling rate may vary for each moving object, we use linear interpolation to temporally align the predicted locations at a common time-slice with a stable sampling (alignment) rate $sr$. 
        
        Given a timeslice $TS_{now}$, EvolvingClusters works in a nutshell, as follows:
        \begin{itemize}
            \item Calculates the pairwise distance for each object within $TS_{now}$, and drop the locations with distance less than $\theta$;
            \item Creates a graph based on the filtered locations, and extract its Maximal Connected Components (MCS) and Cliques (MC) with respect to $c$;
            \item Maintains the currently active (and inactive) clusters, given the MCS and MC of $TS_{now}$ and the recent (active) pattern history; and 
            \item Outputs the eligible active patterns with respect to $c, t$ and $\theta$. 
        \end{itemize}

        The output of EvolvingClusters, and by extension of the whole predictive model, is a tuple of four elements, the set of objects $o_{ids}$ that form an evolving cluster, the starting time $st$, the ending time $et$, and the type $tp$ of the group pattern, respectively. For instance, the final output of the model at the example given at Section \ref{sect:ProblemDefinition} would be a set of 4-element tuples, i.e., $\lbrace (P_2, TS_1, TS_5, 2)$, $(P_3, TS_1, TS_5, 1),  (P_4, TS_1, TS_4, 1), (P_5, TS_1, TS_5, 1) \rbrace$ $\bigcup \lbrace (P_4, TS_1,$ \\ $TS_5, 2), (P_6, TS_5, TS_6, 1) \rbrace$, where $tp = 1(2)$ corresponds to MC (respectively, MCS). We observe that, the first four evolving clusters are maintained exactly as found in the historic dataset. In addition to those, we predict (via the FLP model) the following:
        
        \begin{itemize}
            \item $P_4$ becomes inactive at timeslice $TS_5$, but it remains active as an MCS at timeslice $TS_6$
            \item A new evolving cluster $P_6$ is discovered at timeslice $TS_6$
        \end{itemize}
        
        In the Sections that will follow, we define the evaluation measure we use in order to map, each discovered evolving cluster from the predicted to the respective ones in the actual locations, as well present our preliminary results.

\section{Evaluation Measures}\label{sect:EvaluationMeasures}
    The evaluation of a co-movement pattern prediction approach is not a straightforward task, since we need to define how the error between the predicted and the actual co-movement patterns will be quantified. Intuitively, we try to match each predicted co-movement pattern with the most similar actual one. Towards this direction, we need to define a similarity measure between co-movement patterns. In more detail, we break down this problem into three subproblems, the spatial similarity, the temporal similarity and the membership similarity. 
    Concerning the spatial similarity this defined as follows:

    \begin{equation}
        Sim^{spatial}(C_{pred}, C_{act}) = \frac{MBR(C_{pred}) \bigcap MBR(C_{act})}{MBR(C_{pred}) \bigcup MBR(C_{act})}
    \end{equation}\label{eq:SpatialSimilarity}
    
    \noindent where $MBR(C_{pred})$ ($MBR(C_{act})$) is the Minimum Bounding Rectangle of the predicted co-movement pattern (actual co-movement pattern, respectively). Regarding the temporal similarity:
    
    \begin{equation}
        Sim^{temp}(C_{pred}, C_{act}) = \frac{Interval(C_{pred}) \bigcap Interval(C_{act})}{Interval(C_{pred}) \bigcup Interval(C_{act})}
    \end{equation}\label{eq:TemporalSimilarity}
    
    \noindent where $Interval(C_{pred})$ ($Interval(C_{act})$) is the time interval when the the predicted co-movement pattern was valid (actual co-movement pattern, respectively). As for the membership similarity, we adopt the Jaccard similarity:
    
    \begin{equation}
        Sim^{member}(C_{pred}, C_{act}) = \frac{|C_{pred} \bigcap C_{act}|}{|C_{pred} \bigcup C_{act}|}
    \end{equation}\label{eq:MembershipSimilarity}
    
    Finally, we define the co-movement pattern similarity as:
    
    \begin{equation}
        %
        Sim^{\ast}(C_{pred}, C_{act}) = \left\{
            \begin{array}{ll}
                    \begin{aligned}
                         &\ \lambda_1 \cdot Sim^{spatial}\ + \\
                         &\ \lambda_2 \cdot Sim^{temp}\ + \\ 
                         &\ \lambda_3 \cdot Sim^{member}
                    \end{aligned} & Sim^{temp} > 0 \\
                    &      \\
                  0 & Else \\
            \end{array} 
        \right. 
    \end{equation}\label{eq:OverallSimilarity}

    \noindent where $\lambda_1 + \lambda_2 + \lambda_3 = 1,\ \lambda_i \in \left(0, 1\right),\ i \in \lbrace 1, 2, 3 \rbrace$.

    This further implies that a predicted cluster should be correctly matched with the corresponding actual cluster which is not a straightforward procedure. Our methdology for matching each predicted co-movement pattern $C_{pred}$ with the corresponding actual one $C_{act}$ is depicted in Algorithm \ref{algo:ClusterMatching}.
        
        \begin{algorithm}
            \DontPrintSemicolon
            \KwIn{Evolving Clusters disovered using the predicted $EC_p$; and actual $EC_a$ data-points; Measures' weights $\lambda_i, i \in \lbrace 1, 2, 3 \rbrace$}
            \KwOut{``Matched'' Evolving Clusters $EC_m$}

            $EC_m \gets \lbrace \rbrace$\;
            
            \For{\textbf{predicted pattern} $C_{pred} \in EC_p$}{
                $similarity\_scores \gets \lbrace \rbrace$\;
                $topSim = 0$\;
                \For{\textbf{actual pattern} $C_{act} \in EC_a$}{
                    \textbf{calculate} $Sim^{\ast}(C_{pred}, C_{act})$\;
                    \If{$ Sim^{\ast}(C_{pred}, C_{act}) \geq topSim$}{
                        $topSim = Sim^{\ast}(C_{pred}, C_{act})$\;
                        $match_{best} \gets C_{act}$\;
                    }
                }
                $EC_m \gets EC_m \cup match_{best}$\;
            }  
            \caption{{\sc ClusterMatching}. Matches the predicted with the actual evolving clusters}
            \label{algo:ClusterMatching}
        \end{algorithm}
    
    In more detail, we ``match'' each predicted co-movement pattern $C_{pred}$ with the most similar actually detected pattern $C_{act}$. After all predicted clusters get traversed we end up with $EC_m$ wich holds all the ``matchings'', which subsequently will help us in evaluate the prediction procedure by quantifuing the error between the predicted and the actual co-movement patterns.

\section{Experimental Study}\label{sect:ExperimentalStudy}
    In this section, we evaluate our predictive model on a real-life mobility dataset from the maritime domain, and present our preliminary results.

    \subsection{Experimental Setup}\label{subsect:ExperimentalSetup}
        All algorithms were implemented in Python3 (via Anaconda3\footnote{\url{https://www.anaconda.com/}} virtual environments). The experiments were conducted using Apache Kafka\textsuperscript{\textregistered}~with 1 topic for the transmitted (loaded from a CSV file) and predicted locations, as well as 1 consumer for FLP and evolving cluster discovery, respectively. The machine we used is a single node with 8 CPU cores, 16 GB of RAM and 256 GB of HDD, provided by okeanos-knossos\footnote{\url{https://okeanos-knossos.grnet.gr/home/}}, an IAAS service for the Greek Research and Academic Community. 
    
    \subsection{Dataset}\label{subsect:Dataset}
        It is a well-known fact that sensor-based information is prone to errors due to device malfunctioning. Therefore, a necessary step before any experiment(s) is that of pre-processing. In general, pre-processing of mobility data includes data cleansing (e.g. noise elimination) as well as data transformation (e.g. segmentation, temporal alignment), tasks necessary for whatever analysis is going to follow \cite{DBLP:books/sp/PelekisT14}.
        
        In the experiments that will follow, we use a real-life mobility dataset\footnote{Kindly provided to us by \url{https://www.marinetraffic.com/}{MarineTraffic}.} from the maritime domain. The dataset, as product of our preprocessing pipeline, consists of 148,223 records from 246 fishing vessels organized in 2,089 trajectories moving within in Aegean Sea. The dataset ranges in time and space, as follows:
        
            \begin{itemize}
                \item Temporal range: 2\textsuperscript{nd} June, 2018 -- 31\textsuperscript{st} August, 2018 (approx. 3 months)
                \item Spatial range: longitude in [23.006, 28.996]; latitude in [35.345, 40.999]
            \end{itemize}
        
        During the preprocessing stage, we drop erroneous records (i.e. GPS locations) based on a speed threshold $speed_{max}$ as well as stop points (i.e. locations with speed close to zero); afterwards we organize the cleansed data into trajectories based on their pair-wise temporal difference, given a threshold $dt$. Finally, in order to discover evolving clusters, we need a stable and temporally aligned sampling rate. For the aforementioned dataset, we set the following thresholds: $speed_{max} =  50 knots$, $dt = 30 min.$, and alignment rate equal to $1 min.$
        
        The rationale behind these thresholds stems from the characteristics of the dataset which were unveiled after a statistical analysis of the distribution of the $speed$ and $dt$ between succesive points of the same trajectory. 

    \subsection{Preliminary Results}\label{subsect:PreliminaryResults}
        
        In this section, we evaluate the prediction error of the proposed model with respect to the ``ground truth''. We define as ``ground truth'', the discovered evolving clusters on the actual GPS locations. For the pattern discovery phase, we tune $EvolvingClusters$, using $c = 3$ vessels, $d = 3$ timeslices, and $\theta = 1500$ meters. For the following experimental study, we focus -- without loss of generality -- on the MCS output of \textit{EvolvingClusters} (density-based clusters).
        
\begin{figure}[ht]
            \centering
            \includegraphics[width=.6\columnwidth]{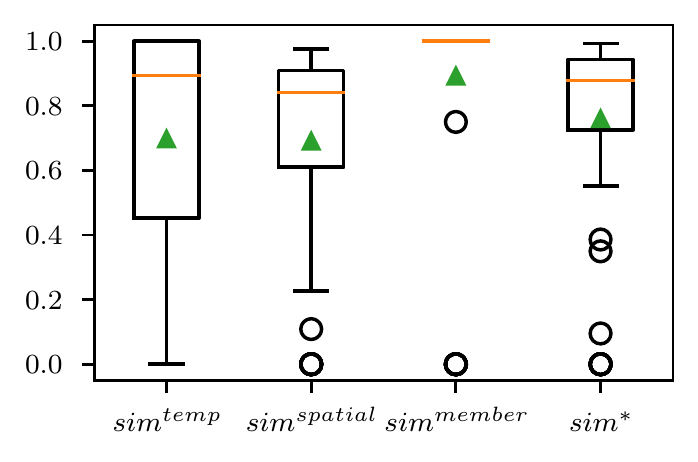}
            \captionsetup{justification=centering}
            \caption{Distribution of Cluster Similarity Measures and Total Cluster Similarity}
            \label{fig:ClusterSimBoxplots}
        \end{figure}

        Figure \ref{fig:ClusterSimBoxplots} illustrates the distribution of the three cluster similarity measures, namely $sim^{temp}$, $sim^{spatial}$, and $sim^{member}$, as well as the overall similarity $Sim^{\ast}$. We observe that the majority of the predicted clusters are very close to their ``ground truth'' values, with the median overall similarity being almost 88\%. This is expected however, as the quality of EvolvingClusters' output is determined by two factors; the selected parameters; and the input data. Focusing on the latter\footnote{The parameter sensitivity of EvolvingClusters is out of the scope of this paper. For more details see \cite{doi:10.1080/13658816.2020.1834562}}, we observe that the algorithm is quite insensitive to prediction errors, as deviations from the actual trajectory has minor impact to $sim^{spatial}$. 
        
\begin{figure}[ht]
            \centering
            \includegraphics[width=.8\columnwidth]{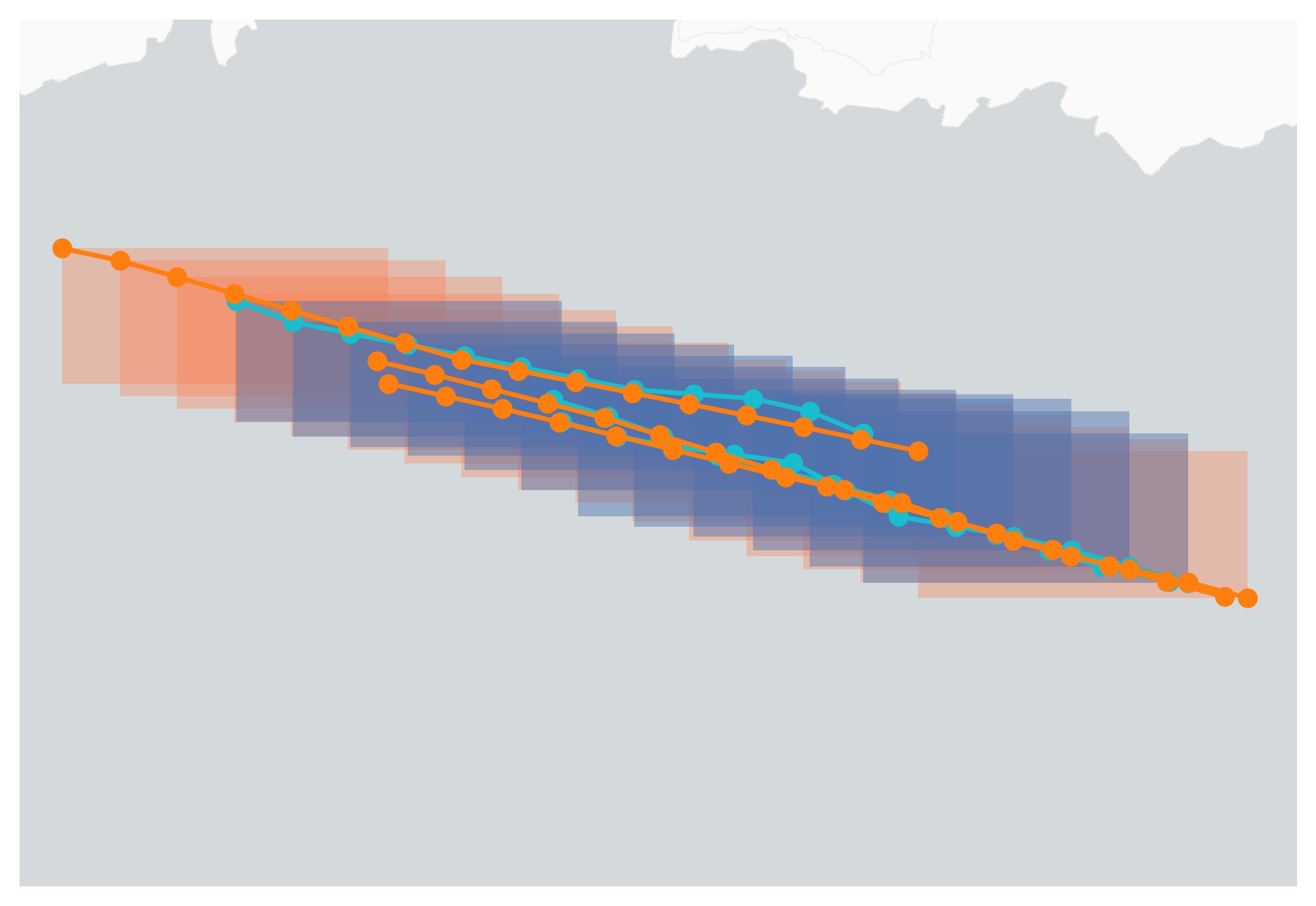}
            \captionsetup{justification=centering}
            \caption{Trajectory of a predicted (blue) vs. an actual evolving cluster (orange)}
            \label{fig:SimMedianMapplot}
        \end{figure}

        Figure \ref{fig:SimMedianMapplot} illustrates the previous discussion. More specifically, for the predicted and corresponding actual MCS with similarity close to the median, we visualize the trajectory of each participating object on the map, as well as the MBRs for each respective timeslice, in order to visualize the clusters' temporal and spatial similarity. It can be observed that deviations from the actual trajectories resulted in minor changes in the area of the points' MBR, and consequently to the overall similarity.
        
        Finally, Table \ref{tab:KafkaConsumerMetrics} presents the metrics on the Kafka Consumers used for the online layer of our predictive model, namely, Record Lag and Consumption Rate. Observing the Record Lag, we deduce that our algorithm can keep up with the data-stream in a timely manner, while looking at Consumption Rate (i.e., the average number of records consumed per second) we conclude that our proposed solution can process up to almost 77 records per second, which is compliant with the online real-time processing scenario.
        
        \begin{table}
            \centering
            \begin{tabular}{lllllll}
                \toprule
                             &    Min.   &    Q25    &    Q50    &     Q75   &     Mean.  &  Max.  \\
                \midrule
                Record Lag   & 0 & 0 & 0 & 0 &  0.01 &  1     \\
                Consump. Rate &   0       &     0     &     0     &      0    &     2.26   &  76.99 \\ 
                
            \end{tabular}
            \caption{Timeliness of the Proposed Methodology using Apache Kafka}
            \label{tab:KafkaConsumerMetrics}
        \end{table}


\section{Conclusions and Future Work}\label{sect:ConclusionsAndFutureWork}
    In this paper, we proposed an accurate solution to the problem of \emph{\prob}, which is divided into two phases: \emph{Future Location Prediction} and \emph{Evolving Cluster Detection}. The proposed method is based on a combination of GRU models and Evolving Cluster Detection algorithm and is evaluated through a real-world dataset from the maritime domain taking into account a novel co-movement pattern similarity measure, which is able to match the predicted clusters with the actual ones. Our study on a real-life maritime dataset demonstrates the efficiency and effectiveness of the proposed methodology. Thus, based on the potential applications, as well as the quality of the results produced, we believe that the proposed model can be a valuable utility for researchers and practitioners alike. In the near future, we aim to develop an online co-movement pattern prediction approach that, instead of breaking the problem at hand into two disjoint sub-problems without any specific synergy (i.e. first predict the future location of objects and then detect future co-movement patterns), will combine the two steps in a unified solution that will be able to directly predict the future co-movement patterns.


\section{Acknowledgements} \label{sec_ack}
This work was partially supported by projects i4Sea (grant T1EDK-03268) and Track\&Know (grant agreement No 780754), which have received funding by the European Regional Development Fund of the EU and Greek national funds (through the Operational Program Competitiveness, Entrepreneurship and Innovation, under the call Research-Create-Innovate) and the EU Horizon 2020 R\&I Programme, respectively.

\bibliographystyle{abbrv}

\bibliography{main}

%
% ...

\end{document}